\documentclass[9pt]{article}
\usepackage{spconf,amsmath,graphicx}
\usepackage{multirow}
\usepackage{diagbox}
\usepackage{amsmath, bm}
\usepackage{comment}
\usepackage{pifont}
\usepackage{bbding}

\usepackage{booktabs}
\usepackage{xcolor}
\usepackage{hyperref}

\title{Mitigate Replication and Copying in Diffusion Models with Generalized Caption and Dual Fusion Enhancement}
\name{Chenghao Li*, Dake Chen*, Yuke Zhang*, Peter A. Beerel\thanks{*Authors contributed equally}}
\address{University of Southern California, Los Angeles, CA}
\begin{document}
%
\maketitle
\begin{abstract}

While diffusion models demonstrate a remarkable capability for generating high-quality images, their tendency to `replicate' training data raises privacy concerns. Although recent research suggests that this replication may stem from the insufficient generalization of training data captions and duplication of training images, effective mitigation strategies remain elusive. To address this gap, our paper first introduces a generality score that measures the caption generality and employ large language model (LLM) to generalize training captions. 
Subsequently, we leverage generalized captions and propose a novel dual fusion enhancement approach to mitigate the replication of diffusion models. Our empirical results demonstrate that our proposed methods can significantly reduce replication by $43.5\%$ compared to the original diffusion model
while maintaining the diversity and quality of generations. Code is available at \href{https://github.com/HowardLi0816/dual-fusion-diffusion}{https://github.com/HowardLi0816/dual-fusion-diffusion}.
\end{abstract}
\section{Introduction}
\label{sec:intro}

The success of deep neural network models has not only fueled the development of Machine Learning Inference as a Service (MLaaS) but has also heightened privacy concerns~\cite{Liu_2023_CVPR, 10247682, Kundu_2023_CVPR, 10323702, Zhang_2023_ICCV, shan2023deep, hwang2023multi, liu2024please}.
While denoising diffusion models excel in generative tasks~\cite{ho2020denoising, rombach2022high, ramesh2022hierarchical, zhang2023adding} and are popular in commercial applications, they face ongoing legal challenges and privacy risks, notably for their tendency to memorize training data, a phenomenon known as \textit{replication}~\cite{somepalli2023diffusion,somepalli2023understanding}. Replication is more likely to appear when fine-tuning a large pre-trained model with a small dataset~\cite{somepalli2023diffusion}.
The common notion attributes the issue of data replication primarily to the lack of general caption of the training images~\cite{somepalli2023understanding}, and the duplication of training images~\cite{vyas2023provable}. 
Specifically, diffusion models tend to memorize images that are either associated with more specific captions~\cite{somepalli2023understanding} or displayed multiple times within the training dataset~\cite{somepalli2023diffusion}. Inspired by these insights, we aim to 
generalize the captions and curtail the use of duplicate images during training to diminish data replication and mitigate the risk of privacy breaches.


The concept of the generality of natural language is abstract and lacks widely accepted quantitative metrics due to the subjective variations among human evaluators. In response, 
we introduce a \textit{generality score} which comprises four  metrics: 
specificity score, broadness score, tense modality score, and abstraction score, 
to quantify the generality of a sentence, and facilitate an analysis of its correlation with data replication. 


Although simplifying captions to a more general form is a direct in approach, it becomes labor-intensive when dealing with a large volume of captions. 
The emergence of Large Language Models (LLMs)~\cite{openai2023gpt4} offers a solution, reducing the need for costly and time-intensive manual processes
across multiple research domains~\cite{xu2023wizardlm, luo2023wizardcoder, mirchandani2023large, qin2023large, yang2023large}. For example, 
WizardLM~\cite{xu2023wizardlm} uses the LLM model, to iteratively regenerate instructions to refine LLaMA~\cite{touvron2023llama}. 
Inspired by this development, we employ LLM to generalize the captions of the diffusion model, achieving a $32.18\%$ reduction in replication compared to the original training captions.

Furthermore, 
we introduce a novel dual fusion enhancement training approach, which involves the stochastic fusion of specific object features with the original image in the latent space, while concurrently fusing the corresponding label embedding with the caption. 
Besides reducing the duplication of training images, this dual fusion process introduces diversity and randomness into the training data, 
effectively alleviating data replication concerns.
Our empirical results demonstrate that the proposed dual fusion approach, in comparison to the original model, can reduce replication by $43.5\%$ while preserving the comparable generation diversity and quality.



We summarize our contributions as follows:
(1) We introduce a generality score combining attributes from four dimensions that measure caption generality.
(2) We generate more general captions using LLM and empirically present the correlation between generality, specificity, and replication.
(3) We propose a novel dual fusion enhancement approach to mitigate replication, and the experimental results demonstrate its effectiveness.






%
%
%

\section{Preliminaries}
\label{sec:Preliminaries}

\subsection{Problem statement}
We denote a text-to-image diffusion model as \emph{Diff}.
The diffusion forward process involves iteratively introducing noise to each training image and transforming it into Gaussian noise. 
The training objective is to predict the distribution of the incorporated noise at different time steps. 
The generative process is expressed as $g_{n} = \mathit{Diff}(p_{n}, r)$, where $g_{n}$ denotes the generated image, $p_{n}$ corresponds to the associated caption, and $r$ represents the initial random noise. In line with Somepalli et al.'s investigation~\cite{somepalli2023understanding}, it has been observed that $X=\{x_{n}\}_{n=1}^N$ and $G=\{g_{n}\}_{n=1}^N$ exhibit a significant degree of memorization, defined as replication. 
Prior work has proposed several defense methods~\cite{somepalli2023understanding}. The Multiple Captions (MC) uses BLIP~\cite{li2022blip} to generate 20 alternative captions for each image and randomly pick one during training. Gaussian Noise (GN) adds Gaussian noise to the text embedding. Random Caption Replacement (RC) substitutes an image's caption with a random sequence of words. Caption Word Repetition (CWR) randomly chooses a word from the provided caption and randomly inserts it at a different position within the same caption.

\subsection{Evaluation of replication}
\label{sec_3_2}

Keeping in line with~\cite{somepalli2023understanding}, we use \textit{replication score}, denoted as $R$, to assess the content replication of the diffusion model \emph{Diff}. Given the original training dataset $X$ and the generated image set $G$, $R$ is defined as the threshold satisfying: $Pr(S < R) = F_{S}(S < R) = 0.95$,
where $S$ is the random variable of the similarity score of two images $x$ and $g$ from $X$ and $G$, respectively. The probability density function of  (PDF) of $S$ is denoted $f_{S}(S)$ and its cumulative distribution function (CDF) is denoted $F_{S}(S<s)$. The similarity score is calculated based on features extracted by the SSCD~\cite{pizzi2022self}, a self-supervised copy detection model, with the official checkpoint. There is a $95\%$ probability that the sample of $S$ will fall below the cutoff point $R$. This ensures that low similarity scores do not distort the overall dataset replication score. 
Specifically, a higher value of $R$ signifies a greater extent of replication.


\section{Strategies for Mitigating Replication in Diffusion Models}
\label{sec:format}


In this section, we introduce 
generality score for captions, 
generating
generalized captions using an LLM, and 
dual fusion enhancement training framework.

\subsection{Semantic generality metrics}
\begin{table}[]
\centering
\resizebox{0.9\columnwidth}{!}{
\begin{tabular}{ccc}

\toprule
Metric &  Equation & Range  \\ \midrule
SI & $1-\frac{Ent+Num}{N_{word}}$ & $[0, 1]$ \\\midrule
BT & $min(\frac{\sum_{i}hypo(noun_{i})}{N_{word}\times 2Avg_{global}}, 1)$  &  $[0,1]$ \\\midrule
TM & $\frac{Count_{\text{pres\&ind}}}{V_{\text{total}}}^*$ & $[0,1]$ \\\midrule
DA & $min(\frac{DA_{caption}}{2DA_{global}}, 1)$ & $[0,1]$\\
\bottomrule
\multicolumn{3}{l}{$*$ TM is set to $0.5$, indicating neutrality, when $V_{total}=0$.}

\end{tabular}}
\caption{Mathematical description of the proposed metrics}
\label{tab:metrics}
\end{table}
Generality is inherently abstract in nature. To quantify it, we introduce four fundamental and intuitive metrics: \textit{specificity of information}, \textit{broadness of terms}, \textit{tense and modality}, and \textit{degree of abstraction}, as mathematically defined in Table~\ref{tab:metrics}. 


\textbf{Specificity of Information (SI)}. 
In essence, a sentence tends to be more general when it contains fewer details. These specific pieces of information are typically presented as named entities such as individual names, locations, specific times, and numerical values~\cite{louis2011general}.
For instance, \textit{"Johnson went to USC at 10 am"} is more specific than \textit{"Students always go to school in the morning"}. The metric computation is shown in Table \ref{tab:metrics} and is based on the number of named entities in a caption $Ent$, the number of numerical details $Num$, and the length of a caption $N_{word}$, all obtained using the spaCy library~\cite{spacy2}.

\textbf{Broadness of Terms (BT)}. 
In hierarchical semantics, broader categories are typically situated higher in the hierarchy and include more specific terms which are referred to as hyponyms~\cite {miller1995wordnet}. For instance, \textit{"animal"} is more general than \textit{"dog"}. We quantify this aspect as in Table \ref{tab:metrics} using the number of hyponyms of each noun $hypo(noun)$ in the caption using WordNet~\cite{miller1995wordnet}.
$Avg_{global}$ is the global average hyponyms count across a lexicon of $30k$ words, ignoring the less-relevant terms~\cite{miller1995wordnet}. 

\textbf{Tense and Modality (TM)}. 
Verbs play a crucial role in determining generality, with sentences in the present tense and indicative mood often being more general. For instance, \textit{"Birds can fly"} is more general than \textit{"The bird was flying"}. 
To quantify this aspect, we calculate the proportion of verbs in the present tense and indicative mood in the sentence using spaCy library~\cite{spacy2}.

\textbf{Degree of Abstraction (DA)}. 
Abstract concepts or words often possess greater generality compared to their concrete counterparts. For example, \textit{“Kindness is a good virtue”} is more general than \textit{“It's good to often greet people with good morning”}. To compute this we use WordNet~\cite{miller1995wordnet} to provide 
a depth measure indicating the degree of abstraction associated with each noun in the caption and compute
the ratio of $DA_{caption}$, the average depth of the synonym sets (synsets) for a caption, to $DA_{global}$, the global average depth.


\textbf{Generality Score (GS)}. 
We aggregate these metrics into a unified measure for each caption by first linearly scaling them into the range $[0,10]$ and then computing their average.

\subsection{Generalize captions}
Creating captions that are general yet convey the semantics of visual images presents a challenge in NLP due to the absence of efficient caption-generalization frameworks. Manual labeling is costly and lacks uniform standards. However, Large Language Models (LLMs) like GPT~\cite{ouyang2022training, bai2022training, touvron2023llama, openai2023gpt4}  have proven to be a cost-effective solution, with their inference capabilities leading to effective text generalization. Therefore, 
we instructed GPT-3.5 to generalize captions at two distinct levels of generality using the following guidelines:

\noindent{\textit{\textbf{General caption}: Convert this caption of an image to a more general caption:[given caption];}}


\noindent{\textit{\textbf{5-word caption}: Make this caption of an image extremely general (result in less than 5 words): [given caption].}}

\noindent{The generalized captions are assessed with the proposed generality score, allowing us to evaluate the effectiveness of the generalization method. Table~\ref{table_1} contrasts SI, BT, TM, DA, and GS values between an original caption and its LLM-generalized versions.}

\begin{table*}[]
\centering
\resizebox{0.8\textwidth}{!}{
\begin{tabular}{c|c|cccccc}
\hline
\textbf{LLM guide} & \textbf{Caption}                                       & \textbf{SI$\uparrow$} & \textbf{BT$\uparrow$} & \textbf{TM$\uparrow$} & \textbf{DA$\uparrow$} & \textbf{GS$\uparrow$} \\ \hline
Original & "2-Ton Multi-Directional Roller Head Pipe Welding Stands"   & 8.18 & 3.67 & 5.0  & 5.15 
& 5.50          \\ \hline
General & "Versatile Pipe Welding Stands for Heavy-Duty Applications" & 10.0 & 3.96 & 10.0 & 5.34 & 7.33          \\ \hline
5-word & "Pipe welding stands in action"                             & 10.0 & 10.0 & 10.0 & 4.61 & \textbf{8.65} \\ \hline
\end{tabular}}
\caption{Example of generalized captions and their generality score}
\label{table_1}
\end{table*}

\begin{figure}[t]
    \centering
    \includegraphics[width=\columnwidth]{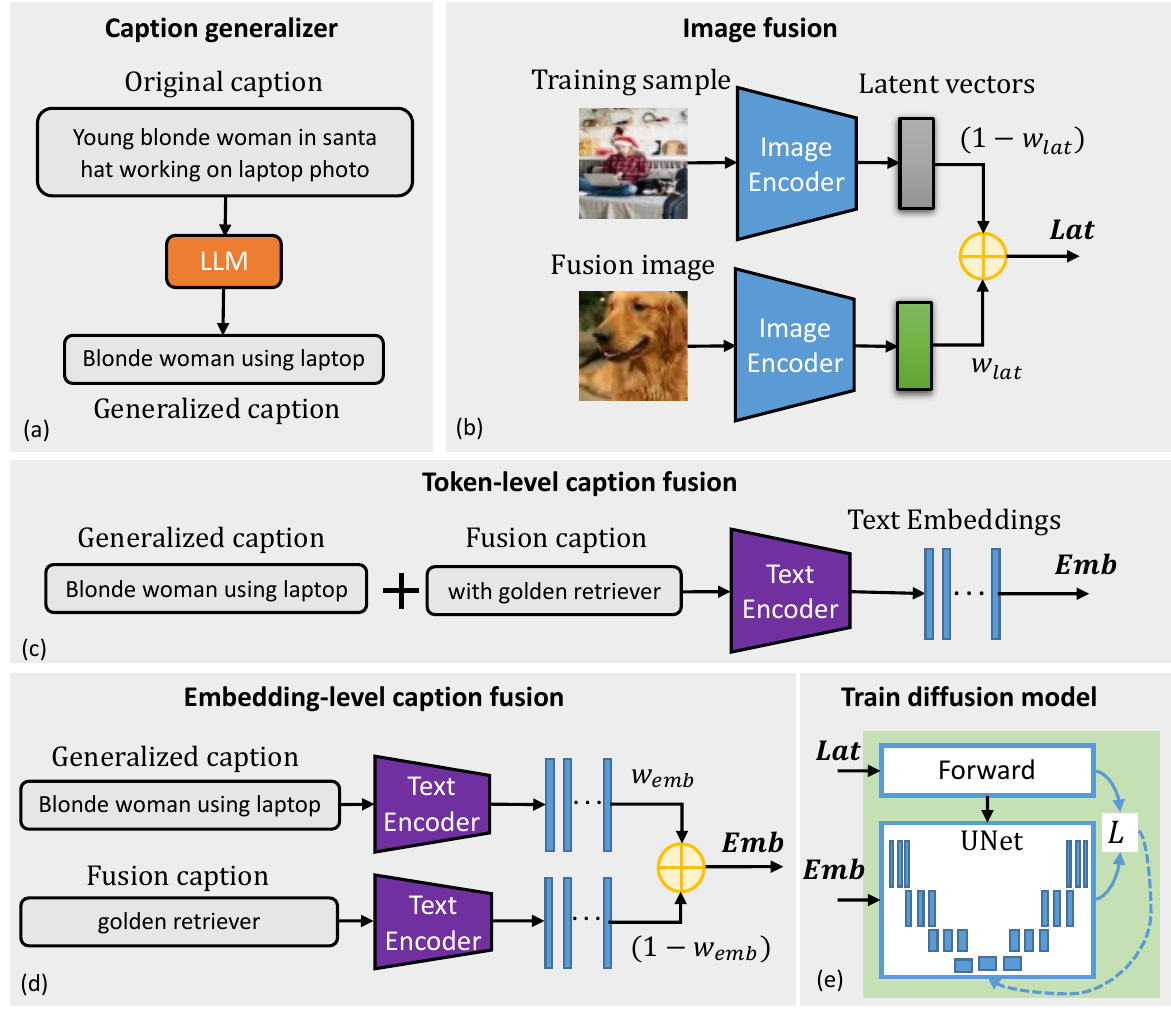}
    \caption{Overview of the proposed methods. (a) Generalize captions with LLM, (b) image fusion, (c) token-level caption fusion, (d) embedding-level caption fusion, (e) train the diffusion model with dual fusion enhancement.}
    \vspace{-3mm}
    \label{fig:overview}
\end{figure}

\subsection{Dual fusion enhancement}
\label{ssec:subhead}
An overview of the proposed dual fusion enhancement is illustrated in Fig.~\ref{fig:overview}. This method utilizes generalized captions to enhance caption generality and incorporates a fusion dataset. This dataset supplies image-text pairs that, when fused with the fine-tuning samples, boost diversity and randomness while minimizing potential image duplication in the fine-tuning dataset.

We define the fine-tuning dataset as $D^{ft}=\{(x_i^{ft}, y_i^{ft} )\}_{i=1}^I$ and the fusion dataset as $D^{fu}=\{(x_j^{fu}, y_j^{fu})\}_{j=1}^J$, where $(x_{i}^{ft}, y_{i}^{ft})$ and $(x_j^{fu}, y_j^{fu})$ are the image-text pair in $D^{ft}$ and $D^{fu}$ respectively.
During the dual fusion training process, both the fine-tuning image $x_{i}^{ft}$ and a randomly selected fusion image  $x_{j}^{fu}$ are passed through a visual encoder, whose function is denoted as $VE$, yielding their respective latent representations. 
Subsequently, we weighted-fuse these two latent representations to obtain an enhanced latent representation, which is expressed as follows:
\begin{equation}
\small
    Lat_{i}=(1-w_{lat})VE(x_i^{ft})+w_{lat}VE(x_{j}^{fu}),
\end{equation}
where $w_{lat}$ is the image latent fusion weight.

Regarding text conditioning, we introduce two fusion modes.
In {\em token-level fusion}, we append the fusion caption $y_{j}^{fu}$ directly to the end of the original caption $y_i^{ft}$. The resulting concatenated caption is then fed into the text encoder, whose function is denoted as $TE$, to obtain the updated text embedding.
In {\em embedding-level fusion}, we separately pass both $y_i^{ft}$ and $y_{j}^{fu}$ through the text encoder, resulting in two distinct text embeddings, which are then weighted-combined at the embedding level to create a unified text representation, represented as follows:
\begin{equation}
\small
    Emb_i=(1-w_{emb})TE(y_i^{ft}) + w_{emb}TE(y_{j}^{fu}),
\end{equation}
where $w_{emb}$ is the text embedding fusion weight.
\noindent
The loss function component for diffusion model is as follows:
\begin{equation}
\small
    L(\theta) = E_{\varepsilon, t}[||\varepsilon - \varepsilon_{\theta}(Lat_{i}^{(t)}, t, Emb_{i})|| ^2],
\end{equation}
where $\varepsilon$ is the noise added during the forward pass, $\varepsilon_{\theta}$ is the predicted noise, and $t$ is the time step.


\section{Experiments}
\label{sec:pagestyle}



\begin{table}[]
\centering
\resizebox{0.98\columnwidth}{!}{
\begin{tabular}{c|ccccccc}
\hline
Method & Baseline & MC~\cite{somepalli2023understanding}    & GN~\cite{somepalli2023understanding}    & RC~\cite{somepalli2023understanding}    & CWR~\cite{somepalli2023understanding}   & \textbf{Ours}  \\ \hline
R$\downarrow$ &0.662     & 0.420 & 0.596 & $\mathit{0.565}$ & $\mathit{0.614}$ & \textbf{0.374} \\ \hline
FID$\downarrow$ &17.394 & 16.831 & 19.504 & - & - & \textbf{15.997} \\\hline
\multicolumn{7}{l}{$*$  The italic replication scores are taken from the paper~\cite{somepalli2023understanding}}
\end{tabular}}

\caption{Comparison of replication mitigation strategies}
\label{table_4}
\end{table}

\noindent \textbf{Experimental setup.}
In line with~\cite {somepalli2023diffusion, somepalli2023understanding}, we focus on reducing replication when fine-tuning a pre-trained diffusion model with small datasets. We fine-tune Stable Diffusion v2.1~\cite{rombach2022high}, pre-rained on LAION~\cite{schuhmann2022laion}, with a subset of LAION-2B~\cite{schuhmann2022laion} with $10k$ randomly selected samples.
We use Tiny-Imagenet~\cite{tiny-imagenet} as the fusion dataset in our dual fusion enhancement, which consists of 200 classes, with 500 images in each class.
We freeze all other modules and only fine-tune the UNet component in the Stable Diffusion. To ensure comparability, we follow the the model hyperparameters in~\cite{somepalli2023understanding}, including $100k$ iterations and a learning rate of $5e^{-6}$. 
We use replication score $R$ 
to evaluate replication on $10k$ generated images, and the Frechet Inception Distance ({\em FID}), which serves as an indicator of the quality and diversity of the model's generated images~\cite{heusel2017gans}.

\noindent \textbf{Caption generality.}
We evaluate the performance of original, general, and 5-word captions in Table~\ref{table_2}. The 5-word captions standout, registering the highest generality score and the lowest replication score, while maintaining a competitive {\em FID}. This highlights the role of caption generality in reducing replication. Thus, we selected the 5-word caption for our subsequent experiments.
\begin{table}[]
\centering
\resizebox{0.8\columnwidth}{!}{
\begin{tabular}{c|c|c|c}
\hline
\diagbox{Metrics}{Caption} & Original & General & \textbf{5-word} \\ \hline
Generality score$\uparrow$        & 5.202              & 5.732               & \textbf{5.846}     \\ \hline
R$\downarrow$        & 0.662                  & 0.529             & \textbf{0.449}         \\ \hline
FID$\downarrow$       & 17.394                 & \textbf{16.837}              & 17.406        \\ \hline

\end{tabular}}
\caption{
Comparison of caption generality levels}
\vspace{-3mm}
\label{table_2}
\end{table}
\noindent \textbf{Comparison with prior-art.}
\begin{figure}[t]
\centering
\includegraphics[scale = 0.38]{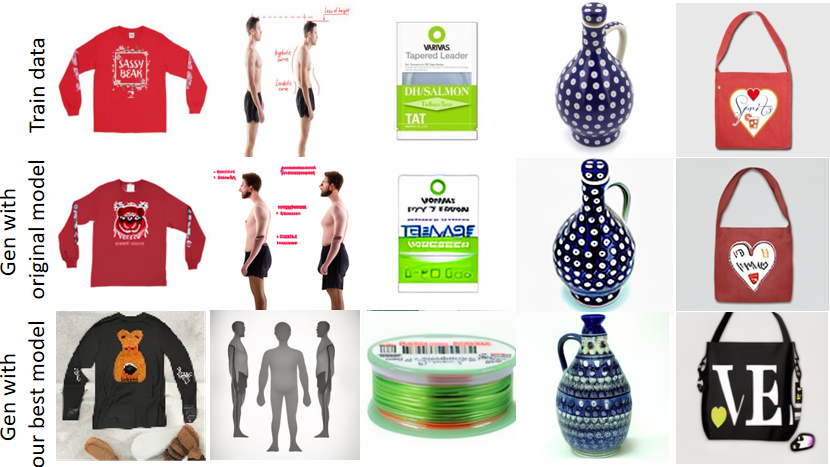}
\caption{Comparison of generated images}
\label{fig:samples}
\end{figure}
In Table~\ref{table_4}, we compare our approach to prior mitigation methods, namely multiple captions (MC), Gaussian noise (GN), random caption replacement (RC), and caption word repetition (CWR)~\cite{somepalli2023understanding}. We use embedding-level fusion with $w_{lat}=0.1$ and $w_{emd}=0.5$ and  
achieves a significant reduction in replication by $43.50\%$ compared to baseline and $10.95\%$ compared to the leading method, MC, with competitive {\em FID}. 



\noindent \textbf{Visualize the generated images.} 
Figure~\ref{fig:samples} showcases five training samples alongside the generated images from both the original diffusion model and our optimized model. Notably, our model considerably diminishes replication, producing images that deviate more from the original training data.

\noindent \textbf{The effects of the fusion weights.}
We report replication scores and {\em FID} values for various fusion weights in Table~\ref{tab:effects_of_w}, covering both token-level and embedding-level scenarios. For token-level dual fusion enhancement, there exists a trade-off between replication score and {\em FID} as $w_{lat}$ increases. This trade-off emerges since a higher $w_{lat}$ allows the fusion image to influence the fine-tuning sample more profoundly, potentially compromising image generation quality. In the embedding-level scenario, using consistent fusion weights for both visual and text elements, we observed a similar balance between replication and {\em FID}. To curtail the influence of the fusion image, we set the latent weight at $0.1$ and progressively adjusted the embedding weight. Our findings suggest that by limiting the dominance of the fusion image and elevating the contribution of the fusion caption, we can strike a balance between R and {\em FID}. This balance is achieved since minimal image fusion can effectively prevent training image duplication, while a greater degree of caption fusion introduces more caption generality.

\begin{table}[t]
\centering
\resizebox{\columnwidth}{!}{
\begin{tabular}{ccc|cccc|cccc}
\hline
\multicolumn{3}{c|}{Token-level} & \multicolumn{4}{c|}{Embedding-level} & \multicolumn{4}{c}{Embedding-level} \\\hline
$w_{lat}$      & R$\downarrow$      & FID$\downarrow$      & $w_{lat}$    & $w_{emb}$    & R$\downarrow$   & FID$\downarrow$   & $w_{lat}$    & $w_{emb}$    & R$\downarrow$   & FID$\downarrow$   \\\hline
0.1         &   0.454     &  \textbf{14.383}        & 0.1       & 0.1       &  0.417   & \textbf{15.680}       & 0.1       & 0.1     &  0.417   & \textbf{15.680}       \\\hline
0.25        &   0.286     &  60.767        & 0.25      & 0.25      &  0.361   &   19.642    & 0.1       & 0.3       & 0.384    &  16.665     \\\hline
0.75         &  \textbf{0.226}      &   73.030       & 0.5       & 0.5       &  \textbf{0.257}   & 61.953      & 0.1       & 0.5       & \textbf{0.374}    & 15.997     \\\hline
\end{tabular}}
\caption{The effects of fusion weights}
\label{tab:effects_of_w}
\end{table}

\noindent \textbf{Ablation study.}
In Table~\ref{tab:ablation}, we present the ablation study of our proposed methods. The 5-word captions yield a notable $32.18\%$ improvement in replication over the baseline. Additionally, the dual fusion enhancement with $w_{lat}=0.1$ and $w_{emb}=0.5$ contributes another $16.7\%$ improvement in replication, all while preserving a competitive {\em FID}.
\begin{table}[h!]
\centering
\resizebox{0.8\columnwidth}{!}{
\begin{tabular}{c|cc|cc}
\hline
\multirow{2}{*}{Model}                                   & \multicolumn{2}{c|}{Ablation}                                                                                      & \multicolumn{2}{c}{Metrics} \\ \cline{2-5} 
                                                         & \begin{tabular}[c]{@{}c@{}}Generalized\\ Caption\end{tabular} & \begin{tabular}[c]{@{}c@{}}Dual \\ Fusion\end{tabular} & Replication     & FID       \\ \hline
Original & \XSolidBrush                                                         & \XSolidBrush                                                      & 0.662           & 17.394    \\ \hline
Ablated  & \XSolidBrush                                                         & \Checkmark                                                      & 0.416           & 16.869    \\ \hline
Ablated & \Checkmark                                                         & \XSolidBrush                                                      & 0.449           & 17.406    \\ \hline
Ours & \Checkmark                                                        & \Checkmark                                                      & 0.374           & 15.997    \\ \hline
\end{tabular}}
\caption{Result of ablation study (\XSolidBrush indicates the module is suppressed and \Checkmark means the module is applied)}
\label{tab:ablation}
\end{table}


\section{Conclusions}
\label{sec:typestyle}
In this paper, we propose a novel approach to bolster the privacy of diffusion models utilizing broader captions and introduce a metric for quantifying caption generality. Our findings underscore that employing generalized captions can substantially reduce data replication of diffusion models. Additionally, our proposed dual fusion enhancement method significantly mitigates data duplication while maintaining the diversity and quality of generations. The extensive experimental results show that our mitigation strategies surpass state-of-the-art methodologies in effectively reducing data replication. In future work, 
we will explore the use of the generality score to guide the caption generalization process and iteratively enhance the generality of each caption.

\vfill\pagebreak



{\footnotesize  
\bibliographystyle{IEEEbib}
\bibliography{strings,refs}
}
\end{document}